\documentclass{article}

\usepackage[preprint]{neurips_2026}

\usepackage[utf8]{inputenc}
\usepackage[T1]{fontenc}
\usepackage{hyperref}
\usepackage{url}
\usepackage{booktabs}
\usepackage{array}
\newcolumntype{L}[1]{>{\raggedright\arraybackslash}p{#1}}
\usepackage{amsfonts}
\usepackage{amsmath}
\usepackage{amssymb}
\usepackage{microtype}
\usepackage{xcolor}
\usepackage{enumitem}
\usepackage{tikz}
\usepackage{algorithm}
\usepackage{algpseudocode}

\usetikzlibrary{arrows.meta,positioning,shapes.geometric,fit,calc,decorations.pathreplacing}

\newcommand{\sys}{\textsc{Grounder}}
\newcommand{\pat}{trusted kernel with a generative shell}

\usepackage{framed}

\title{Never the Number: Structural Abstention for AI\\
       Systems Whose Answers Are Consumed as Fact}

\author{%
  Zhelun (Allen) Wu\thanks{The case study draws on a system the author led
  while at Apple Inc.\ (2021--2024). This report describes architecture and
  design principles only; see the disclosure note in
  Section~\ref{sec:disclosure}. Correspondence:
  \texttt{allenwu94@icloud.com}.}
}

\begin{document}

\maketitle

\begin{abstract}
Large language models have made natural language interfaces to databases newly
credible, but LLM text-to-SQL systems fail in a way that matters for deployment: a
hallucinated column or a mis-aggregated total yields a fluent wrong answer that is
indistinguishable, at the point of use, from a right one. Where the consumer cannot
inspect the generated query, as in enterprise AI deployments and operational
dashboards, and increasingly where the consumer is a tool-using agent rather than a
person, accuracy alone is insufficient: nothing marks which answers to distrust. This
is a reliability problem before it is an accuracy problem.

We propose an architectural pattern for such systems, a \pat{}, resting on one
invariant: a component that can fabricate may influence \emph{which question the
system answers}, never \emph{which value it returns}. A generative shell interprets
underspecified input and phrases replies; a deterministic kernel matches fully
specified questions against a bounded set of answerable question shapes and compiles
them to queries by deterministic execution. The two meet at a confirmation the user reads before any value is
computed, and requests the kernel cannot express are declined rather than
approximated. We call this \emph{structural abstention} and distinguish it from the
\emph{statistical abstention} of selective prediction and calibrated confidence:
refusal here needs no confidence estimate, because unanswerable requests are
unrepresentable.

We specify the pattern implementation-independently, give a five-decision recipe and
work it across three domains, extend the invariant from returned values to the actions
of agentic systems, and report a two-year production case study alongside two generative
alternatives, a fine-tuned parser and a tool-retrieval agent. We close against
enterprise and reliability benchmarks published since, which now measure precisely
the failure the pattern was built to avoid.
\end{abstract}

\vspace{-0.5em}
\vspace{0.5em}

\section{Introduction}
\label{sec:intro}

A regional sales manager wants to know which store in her city sold the most of a
particular product last week. The warehouse holds the answer exactly. She cannot get
at it, because getting at it means knowing which of several tables holds store-level
weekly figures, how the company's fiscal calendar lines up with ordinary dates, and
which of a dozen overlapping store classifications her question implies. So she
writes to an analyst, who is working through a queue, and two days later she gets a
figure by which time the question has answered itself or stopped mattering.

She is not specific to retail. She recurs wherever something needs a value from a
system it cannot query and will act on whatever it is told: a clinician asking how
many beds were free on a ward overnight, a planner asking which components fell below
reorder threshold, a controller asking what a cost centre spent last quarter, and
increasingly a downstream agent acting on a figure no human reviewed at all. What
these share is not subject matter. It is that the answer arrives as an assertion
rather than as a working, and whoever or whatever receives it has no practical means
of auditing it. We call these \emph{fact-consumed} systems, and note that the consumer need not be a
person: a tool-using agent that reads a value and then acts on it is in exactly the
position of the manager who cannot read SQL, with less recourse. The problem is
therefore one of LLM reliability and, in the agentic case, of agent safety, rather
than of model capability.

This class now has a commercial form. Automated data analysis products, variously
marketed as AI analysts or self-serve analytics agents, are fact-consumed systems by
construction: they exist precisely so that someone who cannot write the query does not
have to. The pattern in this report is not a proposal for building such a product. It
is a proposal for the layer such a product needs underneath it if the figures it
returns are to be trusted, and Section~\ref{sec:limitations} is explicit that the
analyst's actual work --- exploration, hypothesis, judgment on novel questions --- is
outside what the pattern can do and should be.

Putting a language model between such a user and the database is the obvious move,
and the difficulty is not that it fails but \emph{how} it fails. A model that
paraphrases a schema fluently will sometimes reference a column that does not exist
or aggregate along the wrong axis \citep{ji2023hallucination}, and one that handles
compositional language will sometimes get the arithmetic wrong
\citep{cobbe2021gsm8k,hendrycks2021math}. Where the analyst reads the query before
trusting it, this is tolerable. Where the query is invisible and the sentence
containing the number is the product, the wrong answers have the same format, the
same confident phrasing, and a plausible magnitude, and the calibration literature
offers limited hope of separating them after the fact \citep{kadavath2022know}. A
system that is right 85\% of the time and cannot indicate which 85\% has not removed
the analyst from the loop; it has removed the analyst and kept the uncertainty.

\begin{figure}[t]
\centering
\begin{tikzpicture}[
  font=\small,
  b/.style={draw, rounded corners=2pt, minimum height=7.5mm, align=center,
            font=\small, inner xsep=3.5pt, fill=white},
  gen/.style={b, fill=black!10},
  arr/.style={-{Stealth[length=1.8mm]}, thick},
  tag/.style={font=\small\itshape, anchor=west},
  reg/.style={draw, rounded corners=3pt, inner sep=2.6mm},
  note/.style={font=\scriptsize, align=center}
]
\node[tag] at (-0.5,1.0) {The usual arrangement};
\node[b]                    (u1) at (0,0) {user};
\node[gen, right=5mm of u1] (m1) {language\\[-2pt]model};
\node[b, right=5mm of m1]   (q1) {query};
\node[b, right=5mm of q1]   (v1) {value};
\node[b, right=5mm of v1]   (a1) {answer};
\foreach \a/\bb in {u1/m1, m1/q1, q1/v1, v1/a1} \draw[arr] (\a) -- (\bb);
\node[reg, dashed, fit=(m1)(a1)] (r1) {};
\node[note, below=1.5mm of r1.south]
  {anything the model gets wrong here arrives as a value the user cannot question};

\begin{scope}[yshift=-3.5cm]
\node[tag] at (-0.5,1.0) {This paper};
\node[b]                    (u2) at (0,0) {user};
\node[gen, right=5mm of u2] (m2) {language\\[-2pt]model};
\node[b, right=5mm of m2]   (c2) {confirmed\\[-2pt]question};
\node[b, right=5mm of c2]   (k2) {kernel};
\node[b, right=5mm of k2]   (q2) {query};
\node[b, right=5mm of q2]   (v2) {value};
\node[gen, right=5mm of v2] (a2) {answer};
\foreach \a/\bb in {u2/m2, m2/c2, c2/k2, k2/q2, q2/v2, v2/a2} \draw[arr] (\a) -- (\bb);
\node[reg, fit=(k2)(v2)] (r2) {};
\node[note, above=1.5mm of r2.north] {no generative component};
\node[note, below=1.5mm of c2.south, xshift=-1mm] {the user reads this\\ and assents};
\node[note, below=1.5mm of a2.south] {phrasing\\ only};
\end{scope}
\end{tikzpicture}
\caption{The core idea. Shaded boxes may fabricate; unshaded boxes may not. In the
usual arrangement a language model produces the query, so any mistake it makes arrives
as a value indistinguishable from a correct one. Here the model works only before the
confirmation and, afterwards, only on wording: the path from confirmed question to
value contains nothing that can invent. The cost is that questions the kernel cannot
express are declined rather than approximated.}
\label{fig:coreidea}
\end{figure}
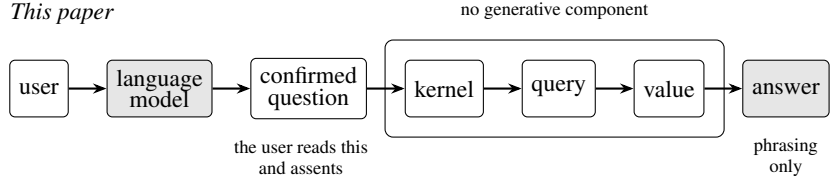

\paragraph{What is new here.}
Two things, and both are conceptual rather than technical. The first is an
architectural invariant, which we state now because the rest of the paper follows
from it.

\begin{framed}
\noindent\textbf{The perimeter invariant.} A component that can fabricate may
influence \emph{which question the system answers}. It may never influence
\emph{which value the system returns}.
\end{framed}

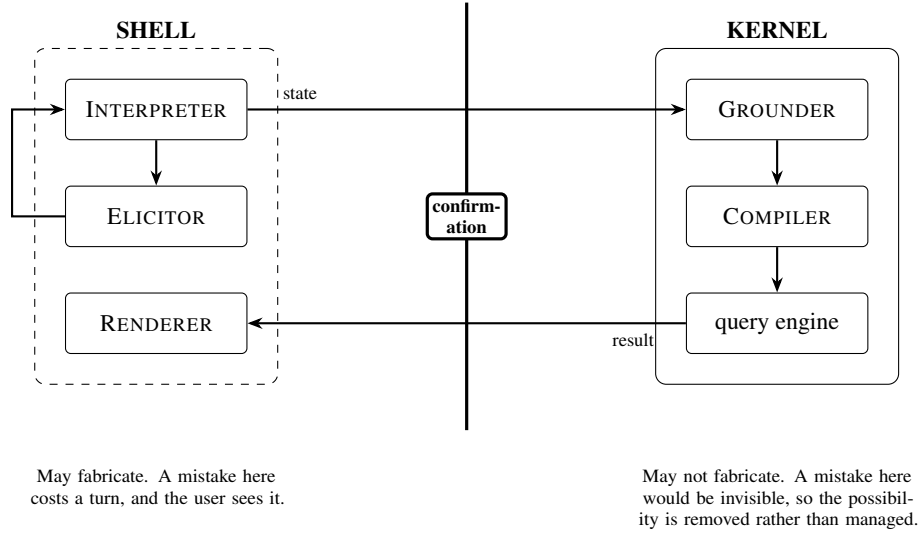
\begin{figure}[t]
\centering
\begin{tikzpicture}[
  font=\small,
  box/.style={draw, rounded corners=2pt, minimum height=8mm, minimum width=24mm,
              align=center, font=\small, fill=white},
  region/.style={draw, rounded corners=4pt, inner sep=4mm},
  arr/.style={-{Stealth[length=2mm]}, thick},
  note/.style={font=\scriptsize, align=center, text width=48mm}
]
\node[box] (interp) {\textsc{Interpreter}};
\node[box, below=6mm of interp] (elic) {\textsc{Elicitor}};
\node[box, below=6mm of elic] (rend) {\textsc{Renderer}};

\node[box, right=58mm of interp] (ground) {\textsc{Grounder}};
\node[box, below=6mm of ground] (comp) {\textsc{Compiler}};
\node[box, below=6mm of comp] (exec) {query engine};

\node[region, dashed, fit=(interp)(elic)(rend),
      label={[font=\bfseries\small]above:SHELL}] (shell) {};
\node[region, fit=(ground)(comp)(exec),
      label={[font=\bfseries\small]above:KERNEL}] (kernel) {};

\draw[arr] (interp) -- (elic);
\draw[arr] (interp.east) -- node[above, font=\scriptsize, pos=0.12] {state} (ground.west);
\draw[arr] (ground) -- (comp);
\draw[arr] (comp) -- (exec);
\draw[arr] (exec.west) -- node[below, font=\scriptsize, pos=0.12] {result} (rend.east);
\draw[arr] (elic.west) -- ++(-7mm,0) |- (interp.west);

\coordinate (top) at ($(shell.north east)!0.5!(kernel.north west)+(0,6mm)$);
\coordinate (bot) at ($(shell.south east)!0.5!(kernel.south west)-(0,6mm)$);
\draw[very thick] (top) -- (bot);
\node[draw, very thick, fill=white, rounded corners=2pt, inner sep=2pt,
      font=\scriptsize\bfseries, align=center]
      at ($(top)!0.5!(bot)$) {confirm-\\ation};

\node[note, below=10mm of shell.south] {May fabricate. A mistake here costs a
turn, and the user sees it.};
\node[note, below=10mm of kernel.south] {May not fabricate. A mistake here would be
invisible, so the possibility is removed rather than managed.};
\end{tikzpicture}
\caption{The perimeter invariant as a structure. A model may operate anywhere in the
shell, where its errors are recoverable and visible. Nothing in the kernel is
generative, because a wrong value there would reach the user unmarked. The two regions
communicate only across the confirmation, which the user reads and assents to before
any value is computed.}
\label{fig:perimeter}
\end{figure}

A model may shape the question and may phrase the reply. Never the number.
Figure~\ref{fig:coreidea} contrasts the resulting arrangement with the usual one, and
Figure~\ref{fig:perimeter} gives its structure. Call the region where the invariant
does not bind the \emph{shell} and the region where it binds absolutely the
\emph{kernel}. The shell interprets underspecified input, decides what to ask next,
and phrases replies; its errors cost a turn and the user sees them. The kernel
matches a fully specified question against the set of questions the system can answer
and turns that question into a query; its errors would be invisible, so the pattern
removes the possibility rather than managing the risk. The two meet at a
\emph{confirmation}: the system states in ordinary language the question it is about
to answer, and the user assents. Everything before that sentence is negotiable;
everything after it is mechanical.

The second contribution is a name for the family this design belongs to.
\emph{Structural abstention} makes unanswerable requests unrepresentable, so that
refusing requires no confidence estimate at all; \emph{statistical abstention}, which
is what the reliability literature has mostly developed, generates a candidate answer
and then estimates whether to trust it
\citep{lee2024trustsql,somov2025confidence,chen2025abstention}. The distinction is
developed in Section~\ref{sec:abstention}. We think it is the more useful of our two
contributions, because it names a design axis that exists independently of anything we
built.

\paragraph{Why this is not one company's engineering problem.}
The invariant says nothing about SQL, and nothing about databases. It says that
whatever produces a consequential value must do so by deterministic execution over a
specification someone authored, and that a model may stand in front of that machinery
but never inside it. Section~\ref{sec:recipe} reduces the pattern to five decisions
and works them for clinical ward reporting and financial reporting alongside our own
enterprise analytics setting, and Section~\ref{sec:actions} extends the invariant from
returned values to executed actions, where it reads: a component that can fabricate
may influence which action is \emph{proposed}, never which action is
\emph{executed}. That version is the one that matters for tool-using agents, and the
failure it prevents is worse there than here, since a wrong figure can be contradicted
later and an executed action has already happened. What the pattern costs is coverage,
and Section~\ref{sec:preconditions} states the three conditions under which that price
is worth paying --- along with the case where it is not.

\paragraph{Structure.}
Section~\ref{sec:related} places the work; Section~\ref{sec:pattern} specifies the
pattern independently of any implementation; Section~\ref{sec:recipe} gives the
instantiation recipe and transfers it; Sections~\ref{sec:arch}--\ref{sec:alternatives}
report a two-year production deployment built on the pattern, together with two
generative alternatives, one abandoned and one that succeeded it and why;
Section~\ref{sec:2026} reassesses the design against work published since; and
Section~\ref{sec:limitations} states what the report does not establish.

\paragraph{Scope and disclosure.}
\label{sec:disclosure}
The case study draws on a production system the author led in industry between 2021
and 2024. This report is limited to architecture and design principles, and excludes
by construction: source code; schemas, table or column names, or any description of a
real data model; datasets; quantitative performance, accuracy, latency, or adoption
metrics; information about customers, users, or personnel; and internal business
processes, product plans, or organizational detail. Figures in examples are invented,
and the schema in Appendix~\ref{app:sql} is synthetic. Deployment is described in the
past tense, since the author's direct knowledge ends in 2024.

\section{Related Work}
\label{sec:related}

\subsection{Natural language interfaces to databases}

The problem \sys{} addresses is old, and so is the observation that
constraining coverage buys reliability. \citet{androutsopoulos1995nlidb} survey
the pre-neural tradition and identify the tension we still face: systems with
broad linguistic coverage are hard to make correct, and systems that are correct
are usually narrow. \citet{popescu2003precise} make the sharpest version of the
argument by defining a class of \emph{semantically tractable} questions ---
those a system can answer with a correctness guarantee --- and having the system
refuse everything else. That is structurally the choice \sys{} makes, with a
template pool playing the role of the tractability criterion.

Later systems added interaction. NaLIR \citep{li2014nalir} presents the user
with its interpretation of a query and lets them correct it before execution,
which is the same insight behind our confirmation turn: showing the user the
question you are about to answer is cheaper than explaining a wrong answer
afterwards. ATHENA \citep{saha2016athena} maps language onto an ontology rather
than directly onto physical tables, decoupling the user's vocabulary from schema
layout --- a separation we implement more crudely, through the template-to-table
mapping described in Section~\ref{sec:templates}. Our contribution relative to
this line is not the constrained-coverage idea, which is theirs, but its
combination with an LLM used strictly outside the correctness-critical path, and
an account of what that combination looks like in production.

\subsection{Neural and LLM text-to-SQL}

The modern line begins with supervised semantic parsing over single tables
\citep{zhong2017seq2sql} and the cross-domain benchmarks that followed
\citep{yu2018spider,li2023bird}. Architectural work improved schema linking
\citep{wang2020ratsql}, and decoding-time constraints improved
well-formedness: PICARD \citep{scholak2021picard} rejects partial outputs that
cannot extend to valid SQL, and LLM-era methods decompose the task into
subproblems with self-correction \citep{pourreza2023dinsql}.

Three things about this literature bear on our design.

First, constrained decoding and our parse-tree compiler attack the same problem
from opposite ends. PICARD guarantees syntactic and schema validity of generated
SQL; it cannot guarantee the query answers the question the user asked. We give up
the ability to generate arbitrary queries in exchange for a guarantee about
semantics: within the pool, the mapping from confirmed question to SQL is fixed
and auditable.

Second, benchmark accuracy on Spider or BIRD is execution accuracy averaged over a
query distribution --- the right metric for a general-purpose parser and the wrong
one for us. A system at 85\% execution accuracy is excellent research and unusable
as an unsupervised oracle, because the 15\% is not labeled. What we needed was not
higher average accuracy but a partition of inputs into answered-correctly and
visibly-declined.

Third, and only visible in hindsight: the benchmarks that made LLM text-to-SQL
look close to solved were not measuring our setting. Two later benchmarks built on
enterprise warehouses report a collapse. On Spider 2.0, whose databases come from
real industrial applications and often exceed a thousand columns, a code-agent
framework over a frontier reasoning model solved 21.3\% of tasks against 91.2\%
on the original Spider \citep{lei2025spider2}. On BEAVER, drawn from private
warehouse query logs, off-the-shelf models scored approximately zero end-to-end,
and supplying the gold tables \emph{and} gold column mappings as hints lifted the
best model only into low single digits \citep{chen2024beaver}. The stated causes
--- schema complexity, business questions requiring multi-table joins and nested
aggregation, and the fact that private schemas are absent from pretraining data
--- are precisely the properties of the warehouse described in
Section~\ref{sec:setting}. Section~\ref{sec:2026} returns to what this means for
the design.

\subsection{Conversational semantic parsing and dialogue state}

SParC \citep{yu2019sparc} and CoSQL \citep{yu2019cosql} extend text-to-SQL to
multi-turn interaction, and CoSQL in particular includes the system-initiated
clarification behavior our guidance mode implements. The difference is
directional: those systems learn when to clarify from data, whereas ours
clarifies when grounding fails, which is a decidable condition. We pay for this
with rigidity and are compensated with a guarantee.

Our dialogue state representation is closer to the task-oriented dialogue
tradition than to semantic parsing. The entity vector is a slot-filling belief
state in the lineage of frame-driven dialogue \citep{bobrow1977gus} and
statistical dialogue state tracking
\citep{henderson2014dstc2,mrksic2017nbt,wu2019trade,budzianowski2018multiwoz};
the decision-theoretic framing of dialogue as action selection over such a state
is reviewed by \citet{young2013pomdp}. Two departures matter. We use a
\emph{pushdown} automaton rather than a flat state machine, because users suspend
a question to ask a sub-question and then resume --- the stack discipline for
discourse segments described by \citet{grosz1986attention}. And our state set is
finite by construction and tied one-to-one to answerable question shapes, so
unlike a learned belief state it cannot represent a configuration for which no
action is defined.

\subsection{Retrieval-augmented generation, tool use, and agentic pipelines}

Our second prototype (Section~\ref{sec:alternatives}) follows what is now called
agentic AI: the retrieval-augmented generation (RAG) \citep{lewis2020rag} and
tool-use \citep{schick2023toolformer,yao2023react} paradigm, in which the model calls
entity-resolution utilities, then composes a query from their returns. Related
in spirit is the observation that offloading computation to an executor rather
than performing it in the model improves arithmetic reliability
\citep{gao2023pal}; interleaved reasoning and acting traces
\citep{yao2023react,wei2022cot} also make the model's path legible, which was
the property we most valued in that prototype. Our reasons for not making it the
primary path were latency and non-determinism rather than accuracy, and we
regard it as the most promising route to extending coverage
(Section~\ref{sec:future}).

\subsection{Reliability, abstention, and knowing when not to answer}
\label{sec:related-abstention}

The concern that motivated this system --- a wrong answer indistinguishable from a
right one --- has since become an explicit research topic, and the framing in that
literature is close enough to ours to be worth stating in its terms. TrustSQL
\citep{lee2024trustsql} identifies two obstacles to deploying text-to-SQL: users
cannot tell which questions the system can correctly answer, and without an
abstention mechanism incorrect SQL goes unnoticed. It scores systems with a
reliability metric that weighs correct answers against wrong ones under a
user-chosen penalty, which is the formal version of the tradeoff
Section~\ref{sec:whyheld} argues informally. A line of work on selective
prediction for SQL generation then attacks the same problem from the model side,
using calibration, entropy, and conformal methods to decide when to abstain
\citep{somov2025confidence,chen2025abstention}.

Our design sits at the structural end of this spectrum. Rather than generating a
query and then estimating whether to trust it, we decline at grounding time,
before any query exists. This trades coverage for a decision that requires no
confidence estimate to be well calibrated. It is a cruder instrument, and it was
available in 2021.

\subsection{Interaction design}

The guidance mechanism is an instance of mixed-initiative interaction
\citep{horvitz1999mixedinit}: the system acts when confident and defers to the
user when uncertain, with the cost of an unnecessary clarification traded
against the cost of a wrong action. Several of the human--AI interaction
guidelines catalogued by \citet{amershi2019guidelines} --- making clear what the
system can do, scoping services to capabilities, supporting efficient correction
--- describe our three guidance modes almost directly, and our \emph{question
suggestion} mode exists precisely to make the capability boundary legible rather
than letting users discover it through wrong answers.

\subsection{Components}

We use an attention mechanism \citep{vaswani2017attention} to score entity
relevance over dialogue context and user profile (Section~\ref{sec:guidance}); a
DSSM-style dual encoder \citep{huang2013dssm} for homepage question
recommendation of answerable questions; an inference-acceleration runtime of the
kind described by \citet{rasley2020deepspeed}; a compact open translation model in
the manner of \citet{tiedemann2020opusmt}; and a neural named-entity recognizer
\citep{lample2016ner,zhang2018lattice} for proper nouns in the users' second
language.

\section{The Pattern}
\label{sec:pattern}

This section states the pattern without reference to any implementation.
Sections~\ref{sec:arch}--\ref{sec:bilingual} then describe one way to build it. A
reader who wants the comparison against end-to-end text-to-SQL and tool-using agents
before the details will find it in Table~\ref{tab:compare}.

\subsection{Component roles and obligations}

The pattern assigns four roles. They need not be four separate processes,
and may be co-located. What matters is that each carries its obligation, and that
the last two discharge theirs without generative components.

\begin{table}[t]
\renewcommand{\arraystretch}{1.15}
\centering
\small
\caption{The four roles. The obligation column is the specification; the
right-hand column states which side of the perimeter invariant each role sits on.}
\label{tab:roles}
\begin{tabular}{@{}L{0.14\textwidth}L{0.25\textwidth}L{0.35\textwidth}L{0.15\textwidth}@{}}
\toprule
Role & Input $\rightarrow$ output & Obligation & Region \\
\midrule
\textsc{Interpreter} &
utterance, state $\rightarrow$ updated state &
Extract whatever is determinate; leave the rest unfilled. Must not invent values
to fill gaps. &
Shell (may be generative) \\ \addlinespace[6pt]

\textsc{Elicitor} &
state $\rightarrow$ utterance &
Choose what to ask next and phrase it. May be wrong; wrongness costs a turn. &
Shell (may be generative) \\ \addlinespace[6pt]

\textsc{Grounder} &
state $\rightarrow$ question, \emph{or} an explicit failure &
Decide whether the state denotes an answerable question. \textbf{Must be total and
decidable}: return a question or a failure, never an approximation. &
Kernel (must not be generative) \\ \addlinespace[6pt]

\textsc{Compiler} &
question $\rightarrow$ query &
Map a grounded question to a query. \textbf{Must be a function}: same question,
same query, always. &
Kernel (must not be generative) \\
\bottomrule
\end{tabular}
\end{table}

Two obligations carry the pattern's guarantees and deserve emphasis.

\textsc{Grounder} must be \emph{total}: for every state it either yields a question
or reports that it cannot. Refusal thereby becomes a first-class
outcome instead of an error path. A generative component cannot discharge this
obligation, and the reason is not inaccuracy: it has no way to signal failure that a
caller can distinguish from an answer.

\textsc{Compiler} must be a \emph{function} in the mathematical sense. Determinism is what makes the system auditable. The
query behind any answer can be reconstructed from the confirmed question alone,
with no need for logs of what a model happened to produce that day.

\subsection{Representing partial specification}

The pattern needs a state that can represent \emph{incompleteness} explicitly, and
this is the requirement that rules out passing raw text between turns. Any
representation works if it supports three operations:

\begin{enumerate}[leftmargin=*,itemsep=1pt]
  \item \textbf{Completeness test.} Given a state, decide whether it denotes an
        answerable question. This is what \textsc{Grounder} needs.
  \item \textbf{Gap enumeration.} Given an incomplete state, enumerate what is
        missing. This is what \textsc{Elicitor} needs to ask a useful question.
  \item \textbf{Consistency test.} Given a state, decide whether it
        over-specifies: whether the user has asked for something the system cannot
        coherently express. This third operation is easy to omit, and
        Section~\ref{sec:anomaly} shows why much of the value lies there.
\end{enumerate}

Section~\ref{sec:entities} gives one concrete representation. The pattern is
indifferent to the choice.

\subsection{The dialogue policy}

The invariant, plus a totality obligation on \textsc{Grounder}, forces a
three-branch policy. There is nothing to design here; it falls out.

\begin{algorithm}[t]
\small
\caption{The bounded-answer dialogue loop. The three branches are exhaustive
because \textsc{Grounder} is total: a state either grounds, or is incomplete, or
is inconsistent.}
\label{alg:loop}
\begin{algorithmic}[1]
\State $\sigma \gets \textsc{InitialState}(\mathit{profile})$
\Loop
  \State $u \gets$ user utterance
  \State $\sigma \gets \textsc{Interpreter}(u, \sigma)$
  \Comment{shell; extracts only what is determinate}
  \If{$\textsc{Inconsistent}(\sigma)$}
    \State emit \textsc{Elicitor}$(\sigma)$ naming the conflict; await resolution
    \Comment{\emph{correction}}
  \ElsIf{$q \gets \textsc{Grounder}(\sigma)$ succeeds}
    \State emit confirmation of $q$; \textbf{await assent}
    \Comment{the perimeter}
    \If{assented}
      \State $r \gets \textsc{Execute}(\textsc{Compiler}(q))$
      \Comment{kernel; deterministic}
      \State emit \textsc{Render}$(q, r)$
      \Comment{shell; phrasing only, over a known result}
    \EndIf
  \ElsIf{$\textsc{Progressing}(\sigma)$}
    \State emit \textsc{Elicitor}$(\sigma)$ asking for a missing element
    \Comment{\emph{elicitation}}
  \Else
    \State emit nearest answerable questions
    \Comment{\emph{disclosure}: state what \emph{can} be asked}
  \EndIf
\EndLoop
\end{algorithmic}
\end{algorithm}

The final branch is the one most systems omit and the one that does the most work.
When elicitation stops converging, the user is asking for something outside the
system's range. The pattern then requires the system to say what it \emph{can}
answer, instead of continuing to interrogate or falling back on a guess. This
turns the capability boundary from something users discover through wrong answers
into something the interface tells them, which is the property
\citet{amershi2019guidelines} identify as making clear what a system can do, and
which \citet{lee2024trustsql} later identify as one of two obstacles to deploying
this class of system at all.

\subsection{What the pattern guarantees}

The guarantees are modest, and stating them precisely is the point.

\begin{enumerate}[leftmargin=*,itemsep=1pt]
  \item \textbf{No fabricated values.} No returned figure was produced by a
        generative component. Computation is delegated to the query engine.
  \item \textbf{Auditability without logging.} Any answer's query is recoverable
        from its confirmed question, because \textsc{Compiler} is a function.
  \item \textbf{Failure is visible.} The residual failure mode is refusal of
        an answerable-in-principle question. That is an error of coverage, which
        the user sees, and not an error of fact, which the user does not.
  \item \textbf{Semantic confirmation.} The user has read the question their value
        answers, in language they can evaluate, before receiving it.
\end{enumerate}

What the pattern does \emph{not} guarantee: that the confirmed question was the
one the user wanted (they may assent carelessly); that the mapping from question
to query is semantically correct, since a mis-authored template is a silent bug.
Correctness there is asserted at authoring time by whoever knew the data model; the
pattern relocates trust to that moment instead of eliminating the need for it.
Nor does it guarantee
or anything at all about coverage.

\subsection{Preconditions, and when not to use this}
\label{sec:preconditions}

The pattern trades coverage for failure visibility. Three conditions determine
whether that trade is favorable.

\textbf{1. Output is consumed as fact.} The user cannot or will not inspect the
generated query. If they can, as with an analyst running a
copilot in a notebook, a generative system is strictly better: the human is the
verifier that the pattern otherwise has to replace with structure.

\textbf{2. Question demand is structurally repetitive.} Questions recur in a small
number of shapes with varying parameters. If every question is novel, a bounded
set of shapes cannot cover enough to be useful, and the pattern degrades into an
elaborate way of saying no.

\textbf{3. Errors are costly relative to refusals.} A wrong figure propagates into
a decision; a refusal costs a delay. In brainstorming, exploration, and low-stakes
summarization, refusal costs more than error and the trade inverts.
\citet{lee2024trustsql} give a way to make this exchange rate explicit as a
penalty weight, which is the natural way to decide the question quantitatively.

If all three hold, the pattern is appropriate and its coverage cost is worth
paying. If the first fails, do not use it. If only the second fails, a hybrid is
indicated (Section~\ref{sec:future}): kernel for what grounds, generative fallback
for what does not, routed on grounding success rather than on a confidence score.

\subsection{LLM reliability: structural versus statistical abstention}
\label{sec:abstention}

Two families of technique prevent a system from answering wrongly, and they are
worth naming, because this report belongs squarely to the second while the
literature has mostly developed the first.

\emph{Statistical abstention} generates a candidate answer and then estimates
whether to trust it, using calibration, sampling agreement, entropy, or conformal
bounds \citep{somov2025confidence,chen2025abstention}. \emph{Structural abstention}
makes unanswerable requests unrepresentable, so that refusal requires no estimate at
all: the system declines because it has nothing to say, not because a score fell
below a threshold.

The pattern is a coarse member of the structural family:
the price of needing no calibrated confidence is that coverage is fixed by however
many question shapes someone has authored. The two families compose, and we regard the
composition as the most promising direction (Section~\ref{sec:future}): structural
refusal for what cannot be expressed, statistical abstention over a generative
fallback for what can be expressed but not grounded.

\section{Instantiating the Pattern}
\label{sec:recipe}

Section~\ref{sec:pattern} specifies roles and obligations but not how to fill them.
This section gives the five decisions an implementation must make, works them across
three unrelated domains, and then generalizes the invariant beyond returned values.
The point of the exercise is to separate what the pattern requires from what our
particular deployment happened to choose; a reader in another domain should be able
to answer the five questions for themselves and stop reading.

\subsection{The five decisions}

\begin{enumerate}[leftmargin=*,itemsep=2pt]
  \item \textbf{What is the unit of answerability?} The kernel needs an enumerable
        set of things it can answer. Ours was a pool of question templates with
        typed slots. Alternatives include a set of parameterized reports, a
        published metric definitions, or a bounded grammar over a semantic model.
        The requirement is only that membership be decidable and the set be
        authored by someone who knows the data.
  \item \textbf{What represents a partial request?} Something that can express
        incompleteness and support the three operations of
        Section~\ref{sec:pattern}: completeness, gap enumeration, and consistency.
        A slot frame is the obvious choice, and the one we made.
  \item \textbf{What makes the request-to-answer mapping trustworthy?} The
        \textsc{Compiler} must be a function, but what it compiles \emph{to} is
        free: SQL text, a call to a governed metrics API, a stored procedure, a
        parameterized report invocation. The more constrained the target, the less
        there is to get wrong.
  \item \textbf{Where is the confirmation, and what does it show?} The user must be
        shown the request in terms they can evaluate. This is a domain question,
        not a technical one: the right rendering is whatever vocabulary the user
        would have used to ask a colleague.
  \item \textbf{What happens on refusal?} The weakest link in most
        implementations. The pattern requires that the system disclose what it
        \emph{can} do; the design choice is whether that is a list of nearest
        answerable requests, a handoff to a human, or a fallback to a generative
        path with the difference made visible.
\end{enumerate}

\subsection{The recipe worked in three domains}

Table~\ref{tab:transfer} answers these questions for our deployment and for two
domains we have not built in. The second and third columns are constructions, not
reports of systems that exist; we include them because a pattern that cannot be
specialized on paper is unlikely to be specializable in practice, and because the
exercise exposes which decisions are genuinely domain-dependent.

Three observations from the exercise.

\begin{table}[h]
\renewcommand{\arraystretch}{1.15}
\centering
\small
\caption{The five decisions, answered for three domains. Rows 1--5 are the decisions
of Section~\ref{sec:recipe}; the final rows record what changes. Only two of the five
decisions turn out to be strongly domain-dependent, which is the substance of the
transferability claim.}
\label{tab:transfer}
\begin{tabular}{@{}L{0.155\textwidth}L{0.255\textwidth}L{0.245\textwidth}L{0.245\textwidth}@{}}
\toprule
Decision & Retail channel analytics \emph{(this report)} & Ward operations reporting & Cost-centre financial reporting \\
\midrule
1. Unit of answerability &
Question templates with typed slots &
Templates over a governed clinical metric set &
Published metric definitions in a finance semantic layer \\ \addlinespace[6pt]

2. Partial request &
Entity vector plus value dictionary &
Slot frame: ward, measure, shift window &
Slot frame: entity, account, period, basis \\ \addlinespace[6pt]

3. Compilation target &
SQL over warehouse relations &
Calls to a governed metrics service &
Parameterized report invocation \\ \addlinespace[6pt]

4. Confirmation surface &
Restated question in sales vocabulary &
Restated in clinical vocabulary, with the shift boundary made explicit &
Restated with the accounting basis and period made explicit \\ \addlinespace[6pt]

5. On refusal &
Nearest answerable questions &
Handoff to an on-call analyst &
Pointer to the report that does cover it \\
\midrule
\emph{Why the invariant matters here} &
A wrong figure enters a commercial decision &
A wrong figure enters a staffing or safety decision &
A wrong figure may enter a filing \\ \addlinespace[6pt]

\emph{Hardest decision} &
1 (authoring cost) &
4 (clinical vocabulary is contested) &
3 (the basis, not the number, is the ambiguity) \\
\bottomrule
\end{tabular}
\end{table}

\textbf{Decisions 2 and 5 barely move.} A slot frame works everywhere, and the
refusal branch differs only in where it points. These are the parts of the pattern
that can be implemented once and reused.

\textbf{Decision 4 is where domain expertise is unavoidable.} What counts as a
legible restatement is a question about the user's professional vocabulary, and
getting it wrong voids the pattern's main guarantee without producing any visible
symptom: users assent to confirmations they have not really understood. In the
financial column the ambiguity is not which number but on what basis it was
computed, so a confirmation that omits the basis is worse than no confirmation, since
it manufactures false assurance.

\textbf{Decision 3 gets easier as the target gets narrower.} We compiled to SQL
because that was what we had. A deployment with a governed metrics service or a
semantic layer available should compile to that instead: the narrower the target
language, the smaller the space of ways a mis-authored mapping can be wrong. This is
the one place where the intervening years have straightforwardly improved on our
implementation.

\subsection{From values to actions: agentic AI and tool-using systems}
\label{sec:actions}

The invariant is stated over returned values because that is what our system
returned. It generalizes without modification to systems that \emph{act}:

\begin{framed}
\noindent A component that can fabricate may influence \emph{which action is
proposed}. It may never influence \emph{which action is executed}.
\end{framed}

The kernel's job becomes mapping a confirmed intent to a parameterized effect
rather than to a query, and the confirmation surface becomes the more familiar
are-you-sure. Everything else carries over: the totality obligation on
\textsc{Grounder} becomes the requirement that unsupported requests be refused
rather than improvised into some adjacent action, and the determinism obligation on
\textsc{Compiler} becomes the requirement that a confirmed intent map to exactly one
effect.

The domains where this matters are not exotic. Any setting that combines a fixed data
model, recurring question shapes, and a reader who acts on the answer without auditing
it is a candidate: enterprise and operational reporting, clinical and ward-level
summaries, financial and regulatory reporting, supply-chain and inventory status, and
the growing class of agents that read a figure and then do something about it. What
varies across them is the confirmation vocabulary and the compilation target
(Table~\ref{tab:transfer}); the invariant does not vary at all.

We flag the action form because the failure mode that motivated the pattern is worse
for actions than for values, which makes it a question of agent safety rather than
only of output quality. A wrong figure can at least be contradicted later by a
right one. An executed action has already happened, and tool-using agents are now
routinely given effects to execute on the strength of an inference no one inspected.
We have not built such a system, and offer the generalization as a hypothesis with a
clear shape rather than as a result.

\section{A Deployed Instantiation}
\label{sec:arch}

We now fill in the first column of Table~\ref{tab:transfer} in detail. \sys{} was a
bilingual conversational analytics interface operated in production for roughly two
years.\footnote{The system's internal name is omitted; we refer to it here as
\sys{}. Nothing in the design depends on the identity of the organization, and no
implementation detail specific to it is disclosed
(Section~\ref{sec:disclosure}).} It is offered as an existence proof and a set of
worked choices, not as the only way to satisfy Section~\ref{sec:pattern}.

\subsection{The setting}
\label{sec:setting}

Users were sales staff responsible for retail channel performance across a large
region. Their questions concentrated on a few recurring shapes --- how much of a
product sold, how much is in stock, what is forecast, how many people entered a
store --- each sliced by product, geography, channel, and time. The questions were
quantitative, tolerance for error was low, and the same shape recurred constantly
with different values filled in.

Against Section~\ref{sec:preconditions}: output was consumed as fact (users could
not read SQL and never saw it), demand was structurally repetitive, and a wrong
figure entered a decision while a refusal cost a delay. All three preconditions
held, which is why we describe this instantiation rather than a different one.

Table~\ref{tab:glossary} lists the structural features of the domain that shaped
the design. It is written in terms of feature classes rather than the
organization's vocabulary, because the features are what transfer; a reader in
supply chain or clinical reporting should recognize their own analogues in the
left-hand column.

\begin{table}[t]
\renewcommand{\arraystretch}{1.15}
\centering
\small
\caption{Structural features of the domain, stated generically. The right-hand
column is what matters: each feature is a reason the mapping from question to query
is not obvious, and therefore a reason to fix that mapping at authoring time rather
than infer it per query.}
\label{tab:glossary}
\begin{tabular}{@{}L{0.21\textwidth}L{0.33\textwidth}L{0.37\textwidth}@{}}
\toprule
Feature & Instance in this domain & Consequence for the design \\
\midrule
\textbf{Hierarchical entity with grain-dependent storage} &
A product family and the specific models within it. &
The level asked about determines which physical relation can answer; the two
grains are stored separately. \\ \addlinespace[6pt]

\textbf{Non-disjoint classification} &
Membership labels on a retail store: a store may hold more than one. &
A request spanning two labels has \emph{no} correct aggregation. This is the
consistency test of Section~\ref{sec:pattern}, and Section~\ref{sec:anomaly}
walks the case. \\ \addlinespace[6pt]

\textbf{Structurally divergent subtypes} &
Tiers of intermediary between manufacturer and retailer. &
The subtype named changes the query's \emph{shape}, not merely a filter value. \\ \addlinespace[6pt]

\textbf{Multiple measures, unevenly available} &
Units sold, units in stock, projected sales, visitor counts. &
Measures live in different places and some are meaningless at some time grains,
so measure and grain cannot be chosen independently. \\ \addlinespace[6pt]

\textbf{Organization-specific calendar} &
A fiscal calendar misaligned with the ordinary one, e.g.\ a label denoting the
third week of a fiscal quarter. &
Users mix fiscal, absolute, and relative dates freely, making date handling a
resolution problem rather than a parsing one. \\
\bottomrule
\end{tabular}
\end{table}

A further property cuts across all of them: the warehouse was organized
hierarchically along every axis at once, so the intersection of axes determined
which relation held the answer. There was no single table to query, and choosing
correctly is most of the work --- exactly the choice later enterprise benchmarks
find models make badly \citep{lei2025spider2,chen2024beaver}.

Finally, entity \emph{names} in this domain were largely proper nouns in a
non-English script denoting individual outlets. They are not translatable, not
reliably segmented by general-purpose tokenizers, and must be resolved against a
registry rather than interpreted (Section~\ref{sec:bilingual}).

\subsection{Requirements}

Five requirements followed, and they map onto the pattern:

\begin{enumerate}[leftmargin=*,itemsep=1pt]
  \item \textbf{Intent identification} from underspecified input. Users arrive with
        a product name, not a question. (\textsc{Interpreter}, \textsc{Elicitor}.)
  \item \textbf{No fabricated entities or figures.} A plausible wrong answer is
        worse than a refusal. (The perimeter invariant.)
  \item \textbf{Exact arithmetic.} Aggregations, comparisons, and rankings computed
        by the query engine, not approximated by a model. (\textsc{Compiler}.)
  \item \textbf{Interactive latency.} Conversational turnaround; per-turn model
        round trips are a budget item, not free.
  \item \textbf{Tolerance of scarce training data.} At the outset there was no
        corpus of question--query pairs for the domain and no traffic from which to
        bootstrap one.
\end{enumerate}

Requirement 5 drove sequencing as much as architecture, and generalizes: a
kernel-first system can be built before any data exists and produces labeled data
as a byproduct of being used, whereas a learned parser cannot. The corpus that
later made a fine-tuned alternative possible (Section~\ref{sec:alternatives}) was
the kernel's output.

\subsection{Pipeline}

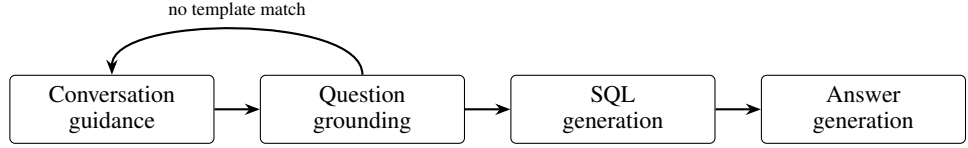
\begin{figure}[t]
\centering
\begin{tikzpicture}[
  node distance=6mm,
  stage/.style={draw, rounded corners=2pt, minimum height=9mm,
                minimum width=27mm, align=center, font=\small},
  arr/.style={-{Stealth[length=2mm]}, thick}
]
  \node[stage] (g)  {Conversation\\guidance};
  \node[stage, right=of g] (q)  {Question\\grounding};
  \node[stage, right=of q] (s)  {SQL\\generation};
  \node[stage, right=of s] (a)  {Answer\\generation};

  \draw[arr] (g) -- (q);
  \draw[arr] (q) -- (s);
  \draw[arr] (s) -- (a);
  \draw[arr] (q.north) .. controls +(0,8mm) and +(0,8mm) ..
             node[above, font=\scriptsize] {no template match} (g.north);
\end{tikzpicture}
\caption{The four-stage pipeline. Only conversation guidance and answer
generation involve a language model; grounding and SQL generation are
deterministic. Failure to match a template returns control to guidance rather
than producing a best-effort query.}
\label{fig:pipeline}
\end{figure}

\sys{} processed each user turn through four stages
(Figure~\ref{fig:pipeline}), which realize the four roles of
Table~\ref{tab:roles}:

\begin{enumerate}[leftmargin=*,itemsep=1pt]
  \item \textbf{Conversation guidance} maintains dialogue state, extracts and
        predicts entities, detects inconsistencies, and decides whether the turn
        has produced an answerable question or requires another exchange.
  \item \textbf{Question grounding} matches the accumulated entity set against
        the template pool and obtains explicit user confirmation of the resolved
        question.
  \item \textbf{SQL generation} compiles the grounded question into a query via
        an intermediate parse tree.
  \item \textbf{Answer generation} renders the result set into a sentence.
\end{enumerate}

Stages 1 and 4 are the shell; stages 2 and 3 are the kernel. A language model
contributes to the shell, where it produces prompts and prose. The kernel ---
the two stages that jointly determine which number the user receives --- contains
no learned component. This is the perimeter invariant of
Section~\ref{sec:intro}, realized as a pipeline boundary, and its consequence is
that the system's errors are errors of coverage (it declines a question it could
in principle have answered) rather than errors of fact.

The return arrow in Figure~\ref{fig:pipeline} is doing more work than it appears
to. In a generative pipeline there is no such arrow: every input produces a query,
because producing a query is what the model does. Here, failure to match a
template is a defined outcome that hands control back to the shell. Most of what
makes the system trustworthy is a consequence of that arrow existing.

\subsection{Entities and entity vectors}
\label{sec:entities}

An \emph{entity} is a typed value the system recognizes: a product, a metric, a
date or range, a geography, a store or reseller. Entities are what carry user
interest, and the system's model of the current dialogue state is essentially a
record of which entity types have been pinned down.

We represent a question as an \emph{entity vector}: a binary vector over
entity-type dimensions indicating which types are present. For the question
\emph{``How many Phone 12 did store A sell yesterday?''} over the dimensions
(product, place, property, date, store), the vector is $(1,0,1,1,1)$ ---
product, property, date and store are supplied; place is
not.\footnote{Place and store are distinct dimensions: a store name may be given
without a geography, and vice versa. The dimensions are not independent, and
Section~\ref{sec:guidance} exploits their correlation for next-entity
prediction.}

This representation is deliberately coarse. It discards which product was named
and retains only that a product was named. That loss is what makes it useful:
the vector is a compact address into the space of partially specified questions,
and the template pool is indexed by the same space. Resolved entity \emph{values}
are carried alongside in a dictionary:

\begin{center}
\small
\begin{tabular}{ll}
\toprule
Key & Value \\
\midrule
\texttt{FAMILY} & phones \\
\texttt{SALES} & sales \\
\texttt{CITY} & Riverton \\
\bottomrule
\end{tabular}
\end{center}

\subsection{Template pool}
\label{sec:templates}

The set of answerable questions is defined extensionally by a pool of templates
with typed slots:

\begin{center}
\small
\begin{tabular}{@{}L{0.52\textwidth}L{0.36\textwidth}@{}}
\toprule
Template & Slot types \\
\midrule
\texttt{How many \_MODEL\_ \_INVENTORY\_ did \_CITY\_ have?}
  & model, measure, geography \\
\texttt{What are the \_FAMILY\_ \_SALES\_ by \_STORECLASS\_ in \_CITY\_?}
  & family, measure, store class, geography \\
\texttt{Which \_STORECLASS\_ are in \_REGION\_?}
  & store class, geography \\
\texttt{How much \_TRAFFIC\_ did \_CITY\_ have in \_QUARTER\_?}
  & measure, geography, fiscal period \\
\bottomrule
\end{tabular}
\end{center}

The pool held a few thousand templates. Each template is associated with the
target table, the slot-to-column mapping, and the aggregation semantics required
to answer it; Appendix~\ref{app:templates} describes the record structure. A
question is \emph{grounded} when its entity vector and dictionary jointly satisfy
a template's slot requirements.

Two properties follow. The pool is an explicit, inspectable specification of
system capability, which means coverage gaps are measurable and the roadmap is a
list. And because grounding is a matching operation rather than a generation
operation, an ungroundable input is detected as such instead of being coerced
into the nearest template.

\section{Conversation Guidance}
\label{sec:guidance}

Conversation guidance converts underspecified input into grounded questions. It
operates in three modes, selected by how much of a template the current turn
satisfies.

\subsection{Three modes}

\paragraph{Question answering.} The input is already complete: it grounds
against a template directly, and the system answers. \emph{User:} ``phone sales
in Riverton.'' \emph{System:} ``Riverton had 5{,}100 phone sales on
2024-01-27.''\footnote{All numeric values in worked examples throughout this
report are synthetic. Question forms, entity types, and system responses are
faithful to the deployed system; sales and inventory figures are not actual
figures and should not be read as such.}
This mode is accurate and fast, and in isolation also brittle: it requires users to
phrase questions the way the pool expects, which is not how people talk.

\paragraph{Question guidance.} The input is incomplete, and the system elicits the
remainder one entity at a time:

\begin{quote}\small
\textbf{User:} sales\\
\textbf{\sys{}:} Got it! Could you provide additional details on sales?\\
\textbf{User:} phones, premium-tier stores, Riverton\\
\textbf{\sys{}:} If I understand correctly, you're asking ``What are the phone
sales by premium-tier store in Riverton?'' --- is that correct?
\end{quote}

The elicitation order is not fixed; it is driven by next-entity prediction
(Section~\ref{sec:prediction}). The mode's weakness is that predictions can be
wrong, producing a clarifying question the user finds irrelevant. The failure is
recoverable, since the user simply supplies a different entity, but it costs a
turn.

\paragraph{Question suggestion.} Neither of the above succeeds: the user's intent
falls outside the pool, or repeated elicitation has failed to converge. Triggered
by explicit dissatisfaction (successive negative responses, or a thumbs-down),
the system abandons the current path and proposes the nearest answerable
questions instead. This is the mode that makes the system's boundary honest. It
does not claim to answer what was asked; it states what can be asked.

\subsection{Automaton over entity states}

Dialogue state is a point $s \in \mathbb{R}^d$ in an embedding space over which a
finite set of discrete \emph{states} is defined, each corresponding to a
configuration of pinned-down entities. Transitions are governed by a pushdown
automaton, whose stack discipline lets the system suspend a partially specified
question, resolve a sub-question, and resume. A user interrupting a sales query to
ask how many stores of a given type exist in the city under discussion is the
common case \citep{grosz1986attention}.

Given the entity vector $v_q \in \mathbb{R}^d$ extracted from the current turn,
the conversation context embedding $e_{ctx} \in \mathbb{R}^m$, and the user
profile embedding $e_u \in \mathbb{R}^n$, the next state is assigned by nearest
neighbor:
\begin{equation}
s_{next} = \mathrm{NearestNeighborState}\bigl(s_{cur} + v_q \otimes (W_a \cdot [e_{ctx}, e_u])\bigr),
\label{eq:transition}
\end{equation}
where $W_a \in \mathbb{R}^{d \times (m+n)}$ is a learned attention matrix
\citep{vaswani2017attention} and $\otimes$ denotes the element-wise product.

Equation~\ref{eq:transition} composes three sources of information. The current
state $s_{cur}$ carries what has been established. The attention term $W_a \cdot
[e_{ctx}, e_u]$ scores entity dimensions by their relevance given the
conversation so far and who is asking. The element-wise product with $v_q$ gates
that scoring by what the current turn actually mentions, so relevance only moves
the state along dimensions the user has touched. Projection back onto the
discrete state set via nearest neighbor is what keeps the dialogue inside the
space of configurations the system can act on.

The discretization is essential rather than incidental. A continuous dialogue state, as in learned belief
trackers \citep{mrksic2017nbt,wu2019trade}, would admit configurations with no
corresponding template and no defined next action. By construction, every state our automaton can occupy has both.

\subsection{Next-entity prediction and prompt generation}
\label{sec:prediction}

The distribution of entity vectors observed in production is far from uniform:
certain entity types co-occur, and a user who has named a product and a metric
is overwhelmingly likely to want a geography next. We exploit this to choose
which entity to elicit, so the system asks the question most likely to complete
the template rather than walking slots in a fixed order.

The user profile embedding $e_u$ personalizes this. Built from the user's query
history, it captures habitual scope: the cities, product families, and store types
this analyst usually asks about. That scope both biases elicitation order and
supplies defaults. A user whose history is dominated by one city, asking about
``phone sales yesterday,'' is more plausibly asking about that city than about
the national total.

Given $s_{cur}$ and the predicted $s_{next}$, the prompt shown to the user is
generated by a language model:
\begin{equation}
\mathit{suggestion} = \mathrm{LLM}(s_{cur}, s_{next}).
\end{equation}
This is the model's entire role in the guidance stage, and it is worth being
precise about why the role is safe. The model does not decide what to ask about;
Equation~\ref{eq:transition} does. The model renders a transition between two known
states into a fluent sentence. If it renders badly the user is confused for a
turn; it cannot cause a wrong number to be returned, because it has no influence
on grounding or query construction.

\subsection{Detecting requests that have no correct answer}
\label{sec:anomaly}

This is the most instructive behavior in the system, and it is worth walking
through slowly, because it is where a bounded pool earns its cost.

Recall from Table~\ref{tab:glossary} that store classes overlap: a single store may
be both premium-tier and standard authorized. Now consider a user who asks for
sales across both classes in one city. The request is grammatical, it is specific,
every term in it is a real term, and a reasonable person would read it as a
perfectly clear question. It has no correct answer. Summing the two classes
double-counts the stores in both; taking either alone answers a different question;
deduplicating silently picks an interpretation the user never chose. No single table
serves the request, and no arithmetic on the available tables is unambiguously what
was meant.

What \sys{} does is notice, and say so:

\begin{quote}\small
\textbf{User:} premium-tier and standard authorized stores, Riverton.\\
\textbf{\sys{}:} You have mentioned premium-tier stores and standard authorized
stores. Let's focus on one at a time. \dots\\
\textbf{User:} premium-tier stores.\\
\textbf{\sys{}:} From what I gather, your question is about ``What are the phone
sales by premium-tier store in Riverton?'' --- is that correct?
\end{quote}

A generative system asked the same thing produces a query. The query is valid, it
runs, and it returns a number of entirely plausible magnitude. Whether that number
double-counts depends on which interpretation the model happened to settle on,
which the user cannot see and would have no reason to suspect. The failure is not
loud. It is a slightly wrong figure in a slide deck, and it survives.

The claim is not that the constrained design handles this case more gracefully. The
claim is that it is \emph{capable of noticing}, because it holds an explicit
representation of what it can and cannot express. Noticing forces the ambiguity
back into the conversation, where the one person who can resolve it is standing:
the person who asked.

Grounding then closes with the confirmation turn, the kernel boundary of
Section~\ref{sec:intro} made visible. The user has read, in ordinary language, the
exact question their number will answer. This is the interaction NaLIR
\citep{li2014nalir} introduced for parse trees, moved up a level to the question
itself, on the theory that users can check a sentence and cannot check a parse.

\subsection{Dialogue history embeddings}

The mechanism above conditions on a context embedding $e_{ctx}$ summarizing the
conversation. A refinement in development at the time of writing maintained a running history
embedding updated per turn through an attention module over the full
question--answer record, rather than primarily the preceding turn. The motivating
case is anaphora, which is also the central difficulty in conversational semantic
parsing benchmarks \citep{yu2019sparc,yu2019cosql}:

\begin{quote}\small
\textbf{User:} phone sales yesterday $\rightarrow$ resolved against profile to a
specific city\\
\textbf{User:} How many premium-tier stores are there?\\
\emph{After attention:} How many premium-tier stores in \{city\}?\\
\textbf{\sys{}:} There are 20 premium-tier stores in \{city\}.\\
\textbf{User:} Which store has the top Phone 12 sales?\\
\textbf{\sys{}:} Are you asking about all stores in \{city\} or these 20
premium-tier stores?
\end{quote}

Attending over the whole history lets ``there'' bind to a city mentioned two turns
earlier. More interestingly, it lets the system recognize that the third question is
ambiguous between two salient scopes, and ask instead of assuming. Open problems include the granularity of the discrete state
embedding and maintaining content diversity in generated prompts across long
sessions. The module was under active development in early 2024 and landed
before the end of the project; the attention-based dialogue engine, pushdown
automaton, and entity-vector extraction described above were all in the shipped
system.

\subsection{Engagement surfaces}
\label{sec:engagement}

Three secondary surfaces reused the same machinery, and are worth recording
because they cost little once the entity representation existed.

\textbf{Homepage question recommendation.} Each user's landing page proposed
questions drawn from the template pool, ranked by a DSSM-style dual encoder
\citep{huang2013dssm} over the profile embedding $e_u$ and the template
representation. Because every recommendation is a pool member, a recommended
question is by construction answerable --- the recommender cannot suggest
something the system will then decline, which is the usual failure of
recommending queries to a natural language interface.

\textbf{Autocomplete ranking.} Partial input was completed against the pool
rather than against a general language model, so completions doubled as
capability disclosure: what the system offers to finish is what it can answer.

\textbf{Daily summaries.} Scheduled digests rendered a user's habitual questions
against fresh data, personalized by the same profile embedding. These required no
dialogue at all --- the questions were already grounded --- which made them the
cheapest surface to build and, per Section~\ref{sec:deployment}, a
disproportionate share of what users valued.

\section{SQL Generation}
\label{sec:sql}

Once a question is grounded, compilation to SQL is deterministic.

\subsection{Matching and parse-tree construction}

Grounding first extracts key information with regular expressions, then selects a
template by combining entity-vector match with sentence similarity against the
pool. The matched template determines an intermediate parse tree whose interior
nodes are SQL constructs --- \texttt{SELECT}, \texttt{FROM}, \texttt{WHERE},
\texttt{GROUP BY}, \texttt{HAVING} --- and comparison and logical operators, and
whose leaves are entity slots. Resolved dictionary values populate the leaves
with concrete table and column names.

For a question of the form \emph{``store Phone 11 sales $\le$ 10 and inventory
$>$ 5 on 2023-09-19,''} the extracted entities (\texttt{STORE}=store,
\texttt{MODEL}=Phone 11, \texttt{MEASURE}=sales, \texttt{MEASURE}=inventory,
\texttt{DATE}=2023-09-19) together with the operators ($\le 10$, $> 5$,
conjunction) yield a tree with an \textsc{AskEntity} root, an \textsc{And} node
over two \textsc{Compare} subtrees, and a \textsc{Where} node carrying the date
predicate. Appendix~\ref{app:sql} works the example through to emitted SQL.

\subsection{Depth-first emission}

SQL text is emitted by depth-first traversal (Algorithm~\ref{alg:dfs}). Each node
type knows how to render itself given its rendered children, which is what makes
the compiler modular: a new aggregate or comparison is a new node type, not a
change to the traversal.

\begin{algorithm}[t]
\small
\caption{Parse tree to SQL by depth-first traversal}
\label{alg:dfs}
\begin{algorithmic}[1]
\Function{Emit}{node, dict}
  \If{\Call{IsLeaf}{node}}
    \State \Return \Call{Resolve}{node.slot, dict} \Comment{table/column/literal}
  \EndIf
  \State $children \gets [\ ]$
  \ForAll{$c \in node.children$}
    \State $children.\mathrm{append}(\Call{Emit}{c, dict})$
  \EndFor
  \State \Return \Call{Render}{node.type, children}
\EndFunction
\end{algorithmic}
\end{algorithm}

The resulting query for the running example targets the store-daily sub-LOB
table, aggregating the sell-out measure with the appropriate predicates and a
\texttt{HAVING} clause carrying the numeric threshold. The same tree structure,
with a different template mapping, compiles against the LOB-grain table when the
question is asked at product-family rather than model level.

Decomposing operations into primitive nodes is what lets the system handle
compositional numeric questions --- thresholds, ratios, top-$k$ rankings ---
without a combinatorial template pool, since a template fixes the question shape
while the tree composes the arithmetic. This is also the component that satisfies
requirement 3 from Section~\ref{sec:setting}: arithmetic is performed by the
database engine, and no model is asked to compute a sum. The same delegation
principle, applied to a generative model rather than a compiler, underlies
program-aided prompting \citep{gao2023pal}.

\subsection{Answer generation}

The result set is rendered into a sentence, using the confirmed question as the
frame. Because the question was assembled from known entities and confirmed by
the user, the rendering is a templating operation over a small result set rather
than a summarization task. Free-text summarization is reserved for the
alternative architecture in Section~\ref{sec:alternatives}, where result shape is
not known in advance.

\section{Bilingual Support}
\label{sec:bilingual}

Users queried in two languages and in mixtures of them: an English-language schema
underneath, and a user population working partly in another language whose entity
names use a different script. Two components handled this, and the division between
them is the transferable part.

The first component translates the shape of the question. A compact open translation
model \citep{tiedemann2020opusmt} normalized each query into the schema's language
before entity extraction, so that a single extraction and grounding path served both
languages instead of two parallel pipelines. A small model was chosen deliberately,
since the translation sits on the critical path of every non-primary-language turn
(requirement 4).

The second component recognizes rather than translates. Translation is the wrong tool
for outlet names, which are proper nouns in a non-Latin script that a general model
will either transliterate inconsistently or render semantically, turning a branch name
into a description of one. We therefore applied a named-entity recognition model
\citep{lample2016ner,zhang2018lattice} to extract outlet and reseller mentions before
translation, and matched them against the canonical registry directly.

The division of labor matters more than either component: the question \emph{form} is
translated, while the entity \emph{values} are recognized and resolved against ground
truth. Translating the whole query and hoping the names survive fails on exactly the
questions users most want to ask, since a specific branch is often the point of the
question.

\section{Designs Considered}
\label{sec:alternatives}

We implemented two generative alternatives. Neither replaced the rule-grounded
core; both informed it. We describe them because the reasons are more instructive
than the conclusion.

\subsection{Fine-tuned text-to-SQL model}

We fine-tuned a customized LLM on three tasks: text-to-SQL, response template
generation, and context completion. Training data came substantially from the
rule-grounded system --- its template pool and the question--SQL pairs accumulated
through its operation --- which is the payoff of having sequenced the heuristic
system first: a few thousand templates and a corpus of question--query pairs on the
order of tens of thousands, all of it a byproduct of the kernel being used. Training
fit on a single multi-GPU node in a few hours, and inference was served through an
acceleration runtime of the kind described by \citet{rasley2020deepspeed}.
Appendix~\ref{app:training} records the procedure.

Iterating on the data taught us more than iterating on the model, and four
interventions account for most of the improvement. Standardizing trigger words across
the training set tightened control over which entities the model would accept and
emit, buying consistency at the cost of tolerance for paraphrase. Shortening the
target queries, by removing redundant qualification and nesting, measurably improved
generation quality, on what we take to be the principle that shorter targets are
easier to learn. Injecting random noise into inputs improved robustness to the
malformed and code-switched queries that characterize real traffic. And appending
manually annotated high-frequency questions improved accuracy on the head of the
query distribution, which is where user-perceived quality is decided.

The model reached usable quality on common question shapes. What it did not
provide was a signal distinguishing its correct outputs from its incorrect ones.
A fluent query against a plausible-looking column, returning a number of the
right magnitude, is indistinguishable at the interface from a correct answer; and
the general difficulty of eliciting reliable self-assessment from language models
\citep{kadavath2022know} gave us little reason to expect a usable confidence
threshold. For our users this is disqualifying as a primary path, though not as a
fallback for inputs the pool cannot ground. In the event we did not deploy it in
either role: the direction that superseded the rule-grounded core was the
tool-retrieval pipeline described next, not this model.

\subsection{Tool-retrieval agent}

The second prototype gives a general LLM a set of retrieval tools and lets it
assemble the query itself, in the manner of retrieval-augmented generation
\citep{lewis2020rag} and learned tool use \citep{schick2023toolformer}. For
\emph{``How many tablets did \{store\} sell three days ago?''} the agent calls an
entity-resolution tool to map the relative date to a fiscal date and the
colloquial store name to a canonical registry entry, then emits SQL against the
resolved values, executes it, and summarizes the result --- using the result's
shape and a few sample rows, with few-shot examples, to produce the final
sentence.

This design has real advantages. It is close to language-agnostic, since entity
resolution rather than translation carries the burden. It handles date formats
the rule-based extractor would need explicit patterns for. And its intermediate
steps are legible: the tool calls and their returns are an audit trail of how the
model arrived at a query \citep{yao2023react}, which is more transparency than
the fine-tuned model offers.

Its cost is latency and non-determinism. Multiple sequential tool calls per turn,
each a model round trip, sit poorly with requirement 4, and the same question can
take different paths on different occasions. Backend work mitigated the first of
these --- an inference-acceleration runtime for the model and a columnar analytical
engine for the queries brought the pipeline inside an interactive budget --- but
mitigation is not elimination.

This is the design the system was transitioning onto at the end of the period
covered by this report. Over the final phase of the project the pipeline moved
from the heuristic question-answering core to a modular LLM pipeline combining
prompt engineering, retrieval augmentation, and tool retrieval, with grounded
answers produced by real-time SQL execution. We describe it here as an
alternative considered because that is what it was for most of the system's life,
and because the comparison in Section~\ref{sec:whyheld} is what determined which
alternative won.

\begin{figure}[t]
\centering
\begin{tikzpicture}[
  font=\small,
  lbl/.style={font=\scriptsize, align=center},
  hdr/.style={font=\small\bfseries},
  arr/.style={-{Stealth[length=2mm]}, thick}
]
\node[hdr] at (0,1.15) {Generative pipeline};
\draw (-3.2,0) rectangle (3.2,0.72);
\draw[fill=black!12] (1.55,0) rectangle (3.2,0.72);
\draw (1.55,0) -- (1.55,0.72);
\node[lbl] at (-0.8,0.36) {answered correctly};
\node[lbl] at (2.38,0.36) {answered\\ \emph{wrongly}};
\node[lbl] at (2.38,-0.55) {indistinguishable, at the point\\ of use, from the
region on the left};
\draw[arr] (2.38,-0.18) -- (2.38,0.02);

\begin{scope}[yshift=-3.1cm]
\node[hdr] at (0,1.15) {Kernel-first pipeline};
\draw (-3.2,0) rectangle (3.2,0.72);
\draw[fill=black!5] (1.05,0) rectangle (3.2,0.72);
\draw (1.05,0) -- (1.05,0.72);
\node[lbl] at (-1.1,0.36) {answered correctly};
\node[lbl] at (2.12,0.36) {declined};
\node[lbl] at (2.12,-0.5) {visible to the user,\\ with a recovery path};
\draw[arr] (2.12,-0.15) -- (2.12,0.02);
\node[lbl, anchor=west] at (-3.15,-0.5) {narrower, by construction};
\end{scope}

\draw[dashed, gray] (1.05,-3.1) -- (1.05,-2.55);
\end{tikzpicture}
\caption{What the pattern trades. The kernel-first system answers a narrower band of
requests and converts the remainder into refusals the user can see and act on; the
generative pipeline answers a wider band but leaves its errors mixed in with its
successes, unmarked. Widths are schematic and carry no measurement: the argument does
not depend on where the boundaries fall, only on the fact that in the upper case there
is no boundary the user can locate.}
\label{fig:partition}
\end{figure}
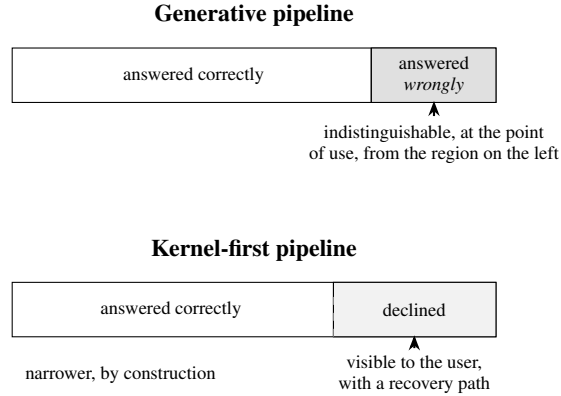

\subsection{What held, and what replaced it}
\label{sec:whyheld}

The comparison isolates a distinction that travels beyond this application, and
Figure~\ref{fig:partition} is the whole of it in one picture.
All three systems make mistakes. They differ in how the mistakes present. The
rule-grounded system fails by declining, visibly, and with a recovery path the
user can act on. A model that generates SQL end to end fails by answering:
fluently, and with no signal that anything went wrong. Where an interface's output
is consumed as fact rather than inspected as a draft, the first failure mode is
preferable even at a substantially lower coverage rate, and this is why the
rule-grounded core carried production traffic for the bulk of the system's life
rather than being replaced as soon as a learned parser became feasible.

How that core was eventually superseded is the more interesting half of the story,
and it supports rather than undercuts the argument. The successor was
not the fine-tuned parser, which had broader coverage and no auditability. It was
the tool-retrieval pipeline, the alternative that retained the two properties the
constrained design existed to protect. Arithmetic stayed delegated to the
database rather than generated, and the tool-call trace preserved a partial audit
path where the fine-tuned model offered none (Table~\ref{tab:compare}). What was
given up was determinism and per-turn latency; what was kept was the refusal to
let a model compute the number.

Table~\ref{tab:compare} summarizes the comparison.

\begin{table}[t]
\centering
\small
\caption{Three architectures for the same task, compared on the axes that decided
between them. The columns correspond to the three approaches available: a bounded
kernel, an end-to-end learned parser, and a tool-using agent. The final row records
what became of each in our deployment.}
\label{tab:compare}
\begin{tabular}{@{}llll@{}}
\toprule
 & Kernel-first & End-to-end & Tool-using \\
 & \emph{(this paper)} & \emph{parser} & \emph{agent} \\
\midrule
Coverage            & Pool-bounded          & Broader             & Broadest \\
Failure mode        & Visible decline       & Fluent wrong answer & Fluent wrong answer \\
Determinism         & Yes                   & Per-checkpoint      & No \\
Latency per turn    & Lowest                & Moderate            & Highest (multi-call) \\
Arithmetic          & Delegated to database & Generated           & Delegated to database \\
Auditability        & Full (fixed mapping)  & None                & Partial (tool trace) \\
Cold-start data     & None required         & Corpus required     & None required \\
\midrule
Disposition         & Production core       & Not deployed        & Successor pipeline \\
\bottomrule
\end{tabular}
\end{table}

\section{Deployment and Usage}
\label{sec:deployment}

\sys{} ran in production for roughly two years, following about a year of development
on the kernel, and established three things. The pattern is implementable: a
kernel-first system with a generative shell can be built and kept running against a
live warehouse by a small team. It is affordable at conversational latency, which was
the requirement most at risk from the generative alternatives
(Section~\ref{sec:alternatives}). And it bootstraps its own successor, since routine
operation accumulated a corpus of question--query pairs large enough to fine-tune a
parser on, which is the concrete form of the sequencing argument in
Section~\ref{sec:setting}.

\sys{} was commissioned because business teams depended on technical staff to query
fragmented dashboards, and after deployment those teams could self-serve the
recurring questions that had previously been requests to an analyst.

There is one measurement we should have taken and did not: the share of turns
resolved by direct answering, by elicitation, and by disclosure. The three modes were
never instrumented separately, and that distribution is the one number that tells a
team whether their pool is converging on demand or falling behind it. Anyone building
to this pattern should instrument the three branches of Algorithm~\ref{alg:loop} from
the first deployment. It is a cheap counter, and it answers the only question that
matters about coverage.


\section{Discussion}
\label{sec:discussion}

Three things about building this surprised us, and all three concern the template
pool rather than the model.

The pool became the specification. We introduced it as an implementation device for
matching questions, and it ended up serving as the system's capability contract:
product discussions about what \sys{} should do next resolved into concrete
additions, and coverage gaps became countable rather than anecdotal. This is an
underrated benefit of defining capability extensionally. A learned parser's coverage
is an empirical question you answer by sampling; a pool's coverage is a list you
read.

Building the deterministic system first was load-bearing rather than expedient. We
initially treated it as scaffolding to be replaced once data existed, but the
causality ran the other way: the system was the only mechanism by which the data
could come to exist (Section~\ref{sec:setting}, requirement 5), and the corpus that
made the fine-tuned prototype possible was its output. Teams in a similar position,
with no question--query corpus and no traffic to bootstrap one, should treat a
constrained system not as a fallback but as a data acquisition strategy.

Data iteration then dominated model iteration. The four interventions of
Section~\ref{sec:alternatives} produced larger quality movements than any
architectural change we tried, which is a familiar lesson, though it was not obvious
in advance that shortening the target queries would matter as much as it did.

Two things we would do differently, both of them cheap and both omitted for no good
reason. We would instrument the three branches of Algorithm~\ref{alg:loop} from the
first deployment rather than retrofitting the measurement
(Section~\ref{sec:deployment}). And we would treat the confirmation turn as a
capability-disclosure surface from the start, since users' mental models of what the
system could answer were formed by trial and error long before the disclosure branch
existed to shape them \citep{amershi2019guidelines}.

\section{Revisiting the Design in 2026}
\label{sec:2026}

Two years have passed since we last worked on this system, and the intervening period
was not a quiet one for the technology it declined to rely on. It would be convenient
to report that the design has aged well. Partly it has, and partly the reasons it has
aged well are not the reasons we expected.

The problem turned out to be harder than the benchmarks of the time suggested. When we
chose a bounded pool over a learned parser in 2021, the honest description of our
reasoning was caution rather than evidence: published text-to-SQL accuracy looked
strong, our schema looked worse than the published ones, and we did not want to find
out the difference in production. Benchmarks built later on enterprise warehouses
suggest the gap was much larger than caution would have guessed. On Spider 2.0, whose
databases come from real industrial applications and often exceed a thousand columns,
an agent framework over a frontier reasoning model solved 21.3\% of tasks where the
same class of system had reached 91.2\% on the original Spider
\citep{lei2025spider2}. On BEAVER, sourced from private warehouse query logs,
off-the-shelf models scored near zero end-to-end, and handing them the correct tables
and column mappings as hints raised the best model only into low single digits
\citep{chen2024beaver}. The failure modes reported are wrong table selection, wrong
column mapping, and missed implicit assumptions, and a template pool resolves all
three by construction, at authoring time, once per question shape. We note this
without triumph: it is retrospective vindication of a decision made for weaker reasons
than the evidence now available would supply.

The gap we designed around has meanwhile become a research area. The report's central
complaint, that a generated query gives no signal distinguishing its right answers
from its wrong ones, is no longer an observation one has to argue for. TrustSQL
\citep{lee2024trustsql} names the same two obstacles we hit --- users cannot tell what
the system can answer, and unnoticed incorrect SQL destroys trust --- and scores
systems by weighing correct answers against wrong ones under an explicit penalty,
which is our qualitative tradeoff made quantitative. Work on selective prediction for
SQL now attacks the same problem from the model side
\citep{somov2025confidence,chen2025abstention}. That literature also locates our
approach in a design space we could not see at the time, as the structural rather than
the statistical answer (Section~\ref{sec:abstention}); we took the structural route
because in 2021 it was the only one that worked, and it remains the route with the
stronger guarantee. The industry pattern that has since converged on it, a curated
semantic layer that a deterministic runtime either serves or errors on, is
recognizably the template pool of Section~\ref{sec:templates} under a better name.

Honesty requires the other column of the ledger, and three parts of the system are now
obsolete. The fine-tuned parser of Section~\ref{sec:alternatives}, with its bespoke
corpus and its from-scratch training run, is not how anyone should approach this now.
The hand-assembled bilingual front end of Section~\ref{sec:bilingual}, a small
translation model plus a purpose-trained entity recognizer, solved a problem that
current multilingual models largely do not have. And the engineering we did to keep a
multi-step tool-using pipeline inside a latency budget has been substantially
commoditized by standardized tool calling and cheaper inference.

None of those obsolescences touch the invariant, which is the part we would keep.
Faster models and standardized tool calls make the shell cheaper to build, and change
nothing about whether a model should be permitted to compute the number a decision
rests on. Reasoning models are better at arithmetic than their predecessors and still
worse at it than a database. The confirmation turn costs a user two seconds and remains
the only mechanism in the design that lets a non-technical person verify what they are
about to be told a number \emph{about}. If we were building this today we would keep
the kernel, discard nearly all of its implementation, and put an agent in the shell,
routing on grounding success instead of on a confidence score
(Section~\ref{sec:future}). The architecture was right and most of the code was
incidental.

\section{Limitations}
\label{sec:limitations}

The central claim of this report has not been tested directly. Testing it would not
mean measuring execution accuracy, which measures the wrong thing here, but measuring
\emph{user-detected} error rate: presenting the same questions to a kernel-first
system and to a generative one, and counting not how often each is wrong but how
often a user notices. We know of no such study, so the claim that visible failure
beats fluent failure remains an argument from design rather than a measured result.
The pattern has also been built once, in one domain, by us; Table~\ref{tab:transfer}
specializes it elsewhere on paper, which shows the five decisions can be answered in
another setting and shows nothing about whether the resulting system would work.

Three costs are intrinsic rather than incidental. Coverage is bounded by
construction, since questions outside the pool can only be redirected, and the pool
grows by manual authoring at a rate set by engineering attention rather than by user
demand. Elicitation costs turns: a user who knows exactly what they want and can
phrase it well is served worse by \sys{} than by a system that simply parses their
sentence, because the design optimizes for the underspecified case and the
confirmation is overhead for anyone who does not need it. And personalization has a
cold start, in that the profile embedding requires history, so new users receive a
generic elicitation order and no defaults.

The sharpest limitation is that a whole class of users is better served by the
architecture this report argues against. Consider an exploratory research analyst
investigating an anomaly in a dataset she knows well. Her questions are novel by
nature, each shaped by the answer to the last; she reads SQL fluently and will
inspect whatever she is given; and being told ``I cannot answer that, but here are
four things I can'' is not a graceful refusal but an obstruction. All three
preconditions of Section~\ref{sec:preconditions} fail at once. For her a generative
copilot is not a compromise but strictly the right tool, and a kernel would make her
slower at no benefit. The pattern earns its cost only where the user cannot verify and
the question shapes repeat, which excludes a large and visible share of the people
currently building natural-language data tools.

It follows that this is not an automated analyst, and it is worth saying so plainly,
because the phrase attaches easily to systems like it. What the pattern automates is
the \emph{queue}, not the judgment. The manager of Section~\ref{sec:intro} waited two
days for a figure that was recurring, parameterized, and known to be answerable; a
human was in that loop only because no machine could be trusted in it. Removing the
recurring requests is what frees an analyst for the exploratory questions no kernel
can express --- which is the same division the pattern draws internally, deterministic
machinery for what is known and a generative shell for what is ambiguous, applied at
the scale of a team rather than a system. A product that claimed to replace the analyst
outright would be claiming to have solved the half of the problem this pattern
deliberately declines.

\section{Directions We Would Have Pursued}
\label{sec:future}

The roadmap below is the one we held at the end of our involvement; we offer it
as design analysis rather than as a statement of anyone's current plans.

The natural extension is a hybrid in which the rule-grounded path handles
groundable questions and a tool-retrieval agent handles the remainder, with the
routing decision made by grounding success rather than by a confidence score.
This preserves the property we care about --- that the deterministic path is
deterministic --- while converting \emph{question suggestion} from a graceful
decline into an actual answer. The open problem is presentation: if answers
arrive from two paths with different reliability, the interface must communicate
that difference without eroding trust in either.

Beyond this: broadening the engagement surfaces of
Section~\ref{sec:engagement}, where the strongest untried idea was to let
observed acceptance of recommended questions drive template authoring, closing
the loop between what users ask for and what the pool covers; and extending the
pool beyond the single retail channel it served to the others alongside it.

\section{Conclusion}

A fact-consumed system has a problem that accuracy alone does not solve: whoever
receives its answers cannot tell the right ones from the wrong ones. The pattern in this report
answers that by dividing the system in two and letting a generative model operate
only on the side where mistakes are visible. A language model shapes the question;
it never computes the answer. The two halves meet at a sentence the user reads and
assents to, and requests the deterministic half cannot express are declined instead
of approximated.

The specific system we built to this design is now largely obsolete in its
implementation and was never the interesting part. What survives is the
invariant, the five decisions of Section~\ref{sec:recipe}, and one observation from
watching the design compete against its own alternatives. When the constrained core
was eventually superseded, the successor was not the alternative with the broadest
coverage. It was the one that kept computation out of the model and left a trace of
how it reached an answer, which are the two properties the kernel existed to
protect. The parser we abandoned was not abandoned for being inaccurate. It was
abandoned because when it was inaccurate, nobody could tell.

\bibliographystyle{plainnat}
\bibliography{refs}

\appendix

\section{Entity Type Inventory}
\label{app:taxonomy}

The types below are those exercised by the examples in this report, named as they
appear in Table~\ref{tab:glossary} rather than in the deployed system, whose slot
names were tied to internal schema vocabulary. The deployed registry was larger.

\begin{table}[h]
\centering
\small
\begin{tabular}{@{}lll@{}}
\toprule
Type & Example values & Notes \\
\midrule
\texttt{FAMILY}     & product family              & Top-level product axis \\
\texttt{MODEL}      & specific model              & Determines which table grain applies \\
\texttt{STORECLASS} & premium-tier, standard, etc. & Overlapping; see \S\ref{sec:anomaly} \\
\texttt{STORE}      & store or reseller           & Reseller tiers are stored separately \\
\texttt{CITY}       & city name                   & \\
\texttt{REGION}     & province or region          & \\
\texttt{SALES}      & sell-out measure            & \\
\texttt{INVENTORY}  & inventory measure           & \\
\texttt{TRAFFIC}    & store visitors              & \\
\texttt{DATE}       & date, \texttt{FY23Q4W3}, ``yesterday'' & Fiscal and relative forms \\
\texttt{QUARTER}    & fiscal quarter              & \\
\texttt{NUM}        & numeric threshold           & Operand for \textsc{Compare} nodes \\
\bottomrule
\end{tabular}
\end{table}

\section{Template Pool Record Structure}
\label{app:templates}

Each template record carries five things: the surface form with typed slots; the
entity vector the form requires, which is what makes matching a lookup rather
than a search; the target table, selected by the grain the slots imply; the
slot-to-column mapping; and the aggregation semantics, which determine what the
compiler of Section~\ref{sec:sql} emits for the \texttt{SELECT} and
\texttt{GROUP BY} clauses.

Authoring a template therefore means answering the question ``which table, at
which grain, aggregated how'' once, at authoring time, rather than at query time.
This is where the correctness guarantee actually lives: the compiler is trusted
because the mapping it applies was written and reviewed by someone who knew the
schema.

The pool was heavily concentrated on a few families, mirroring a query distribution
with a short head and a long thin tail. This is what makes the authoring cost
tractable: the first few hundred templates buy most of the coverage.

\section{Worked SQL Compilation Example}
\label{app:sql}

The schema below is invented for this report. It is not the data model of any
deployed system, and no real table or column name appears; what is faithful is the
\emph{shape} of the compilation --- which node types exist, what each contributes,
and where in the emitted query each lands. That shape is the transferable content.
Figures are likewise invented.

\paragraph{1. Matched question.} \emph{``store Phone 11 sales $\le$ 10 and
inventory $>$ 5 on 2023-09-19''}

\paragraph{2. Entity dictionary.}

\begin{center}
\small
\begin{tabular}{@{}lll@{}}
\toprule
Slot & Value & Resolves to \\
\midrule
\texttt{STORE}     & store        & grouping column \\
\texttt{MODEL}     & Phone 11    & model-level filter \\
\texttt{MEASURE}   & sales        & sell-out measure \\
\texttt{MEASURE}   & inventory    & inventory measure \\
\texttt{DATE}      & 2023-09-19   & fiscal date filter \\
\texttt{NUM}       & 10, 5        & thresholds \\
\bottomrule
\end{tabular}
\end{center}

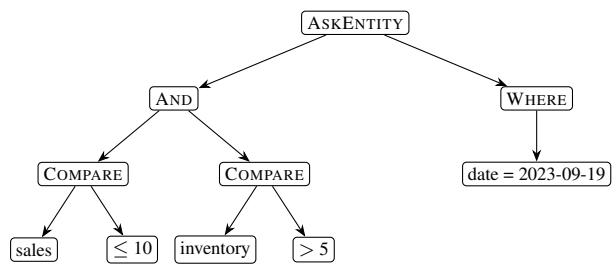
\begin{figure}[h]
\centering
\begin{tikzpicture}[
  level distance=10mm,
  level 1/.style={sibling distance=48mm},
  level 2/.style={sibling distance=24mm},
  level 3/.style={sibling distance=13mm},
  every node/.style={font=\scriptsize, draw, rounded corners=1.5pt, inner sep=2pt},
  edge from parent/.style={draw,-{Stealth[length=1.6mm]}}
]
\node {\textsc{AskEntity}}
  child {node {\textsc{And}}
    child {node {\textsc{Compare}}
      child {node {sales}}
      child {node {$\le$ 10}}}
    child {node {\textsc{Compare}}
      child {node {inventory}}
      child {node {$>$ 5}}}}
  child {node {\textsc{Where}}
    child {node {date = 2023-09-19}}};
\end{tikzpicture}
\caption{The intermediate parse tree for the worked example. Interior nodes are query
constructs and operators; leaves are resolved slot values. Emission is a depth-first
traversal (Algorithm~\ref{alg:dfs}), and each node type renders itself from its
rendered children, which is what lets a new aggregate be added as a node type rather
than as a change to the traversal.}
\label{fig:parsetree}
\end{figure}

\paragraph{3. Parse tree.} Figure~\ref{fig:parsetree}.

\paragraph{4. Emitted SQL.} Depth-first traversal (Algorithm~\ref{alg:dfs})
renders the leaves, then each interior node from its children:

\begin{quote}\small\ttfamily
SELECT store\_name, SUM(sales\_measure) AS agg\_sales \\
FROM \ \ \ store\_sublob\_daily \\
WHERE \ model\_name = 'Phone 11' \\
\ \ AND \ fiscal\_date = '2023-09-19' \\
GROUP BY store\_name \\
HAVING agg\_sales <= 10 AND SUM(inventory\_measure) > 5
\end{quote}

Two details are worth drawing out. The thresholds land in \texttt{HAVING} rather
than \texttt{WHERE} because they constrain aggregates, and it is the
\textsc{Compare} node's renderer --- not the template --- that knows this; a
template author cannot get it wrong. And asking the same question at
product-family rather than model grain changes only the template's target table,
leaving the tree identical.

\section{Fine-Tuning Procedure for the Abandoned Alternative}
\label{app:training}

Recorded as procedure, since the method and its ordering are what transfer.

\paragraph{Where the data came from.} The corpus was not collected for the purpose.
It accumulated as a byproduct of the kernel running in production: every grounded
question paired with the query the compiler deterministically produced for it is a
labeled example, generated at no annotation cost and guaranteed correct by
construction --- the label is not a human's guess at the right query, it \emph{is}
the query that ran. This is the practical payoff of the sequencing argument in
Section~\ref{sec:setting}, and the reason a kernel-first system is a data
acquisition strategy rather than merely a stopgap.

\paragraph{Tasks.} Three, trained jointly: question-to-query generation; generation
of response templates; and completion of context across turns.

\paragraph{Scale and cost.} A few thousand question shapes and a corpus of pairs on
the order of tens of thousands; training fit on a single multi-GPU node in a few
hours. The point worth carrying is the ratio rather than the absolute figures: the
fine-tuning run was cheap next to the year of kernel development that produced its
training data. Anyone reading this as a build-versus-train decision should price the
data, not the GPUs.

\paragraph{What moved quality.} The four data interventions in
Section~\ref{sec:alternatives} --- standardizing trigger vocabulary, shortening
target queries, injecting input noise, and adding annotated head questions ---
produced larger improvements than any architectural change attempted. The
counterintuitive one was shortening the targets: removing redundant qualification
and nesting from the training queries improved generation more than we expected,
which we read as targets being easier to learn when they are closer to minimal.

\paragraph{Why it was not deployed.} Covered in Section~\ref{sec:alternatives}: it
had no way to signal which of its outputs to distrust, and so could not discharge
the \textsc{Grounder} obligation of Table~\ref{tab:roles} regardless of its accuracy.

\end{document}